\documentclass[english,runningheads,a4paper]{llncs}
\usepackage{amssymb}
\usepackage{graphicx}
\usepackage{url}
\usepackage{multirow}
\usepackage{float}
\usepackage{tabularx}
\usepackage{makecell}
\usepackage{subcaption}
\usepackage{tabulary}
\usepackage{hyperref}
\usepackage{hhline}
\usepackage{array}
\newcolumntype{?}{!{\vrule width 1pt}}
\usepackage{makecell}
\newcommand\lword[1]{\leavevmode\nobreak\hskip0pt plus\linewidth\penalty50\hskip0pt plus-\linewidth\nobreak{#1}}

\newcommand{\keywords}[1]{\par\addvspace\baselineskip
\newcommand\footnoteref[1]{\protected@xdef\@thefnmark{\ref{#1}}\@footnotemark}
\noindent\keywordname\enspace\ignorespaces#1}

\newcolumntype{C}[1]{>{\centering\arraybackslash}p{#1}}
\newcolumntype{M}[1]{>{\centering\arraybackslash}m{#1}}
\newcolumntype{N}{@{\arraybackslash}m{0mm}}

\usepackage{color}
\usepackage[usenames,dvipsnames,svgnames,table]{xcolor}

\usepackage{soul}

\newcommand{\com}[1] {}

\usepackage{tikz}
\usetikzlibrary{shapes.geometric}
\usetikzlibrary{calc}
\newcommand{\toppage}{
   \begin{tikzpicture}[remember picture,overlay]
     \node[align=left, anchor=north west]
       at ($(current page.north west) + (1,-1)$)
       {The paper was presented in the MICCAI-LABELS 2018. \url{https://labels.tue-image.nl/previous-editions/labels-2018/}};
   \end{tikzpicture}
}

\begin{document}

\mainmatter

\title{MVOR: A Multi-view RGB-D Operating Room Dataset for 2D and 3D Human Pose Estimation}
\titlerunning{MVOR: A Multi-view RGB-D Operating Room Dataset}

\author{Vinkle Srivastav\inst{1}
\and Thibaut Issenhuth\inst{1}
\and Abdolrahim Kadkhodamohammadi\inst{1}
\and Michel de Mathelin\inst{1}
\and Afshin Gangi\inst{1,2}
\and \\Nicolas Padoy\inst{1}}

\authorrunning{Srivastav et al.}

\urldef{\mailsa}\path{{srivastav | padoy}@unistra.fr}
\institute{ICube, University of Strasbourg, CNRS, IHU Strasbourg, France\\ \mailsa
 \and Radiology Department, University Hospital of Strasbourg, France}

\toctitle{UPDATE}
\tocauthor{UPDATE}
\maketitle

\begin{abstract}
Person detection and pose estimation is a key requirement to develop intelligent context-aware assistance systems. To foster the development of human pose estimation methods and their applications in the Operating Room (OR), we release the Multi-View Operating Room (MVOR) dataset, the first public dataset recorded during real clinical interventions. It consists of 732 synchronized multi-view frames recorded by three RGB-D cameras in a hybrid OR. It also includes the visual challenges present in such environments, such as occlusions and clutter. We provide camera calibration parameters, color and depth frames, human bounding boxes, and 2D/3D pose annotations.  
In this paper, we present the dataset, its annotations, as well as baseline results from several recent person detection and 2D/3D pose estimation methods. Since we need to blur some parts of the images to hide identity and nudity in the released dataset, we also present a comparative study of how the baselines have been impacted by the blurring. Results show a large margin for improvement and suggest that the MVOR dataset\footnote[1]{The MVOR dataset is available at \url{http://camma.u-strasbg.fr/datasets}} can be useful to compare the performance of the different methods.  
\toppage{}

\end{abstract}


\section{Introduction}
\begin{figure}[tb]
\includegraphics[width=0.32\linewidth]{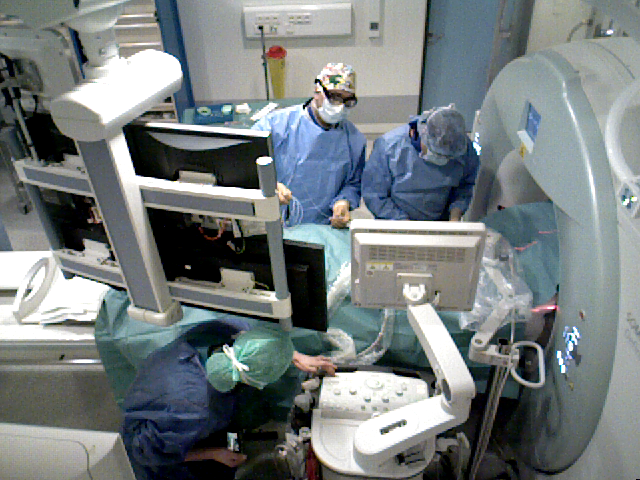} 
\includegraphics[width=0.32\linewidth]{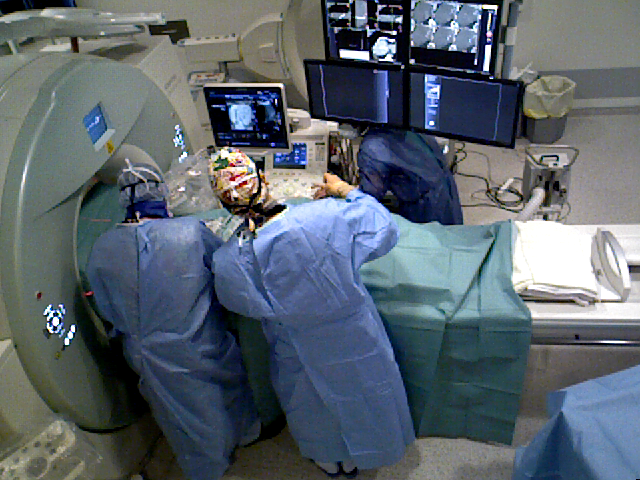} 
\includegraphics[width=0.32\linewidth]{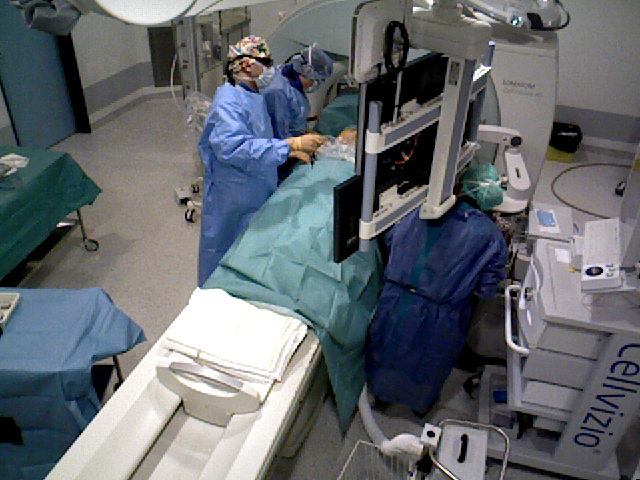} 
\caption{Multi-view images captured by 3 RGBD cameras during a live intervention, illustrating the high visual complexity of an operating room. Three persons are visible in the scene, besides the patient. }
\label{fig:multi-cam-challenges}
\end{figure}

2D and 3D multi-person pose estimation methods have been intensively researched in computer vision for a few decades. They are needed to give machines a better understanding of human activities and thereby enable the analysis of complex visual content and the development of more sophisticated human-computer interactions. Detecting and localizing parts of an articulated object is a difficult task because it requires to capture both part appearances and the structure of the object. Although the rise of deep learning has lead to large improvements, there are still challenging situations, such as crowded scenes with occluded and interleaved keypoints. 
Modern operating rooms present such scenarios with high visual complexity due to occlusion and clutter from various equipment, loose clothes worn by clinicians, and the proximity of the persons present in the scene, as illustrated in Fig.  \ref{fig:multi-cam-challenges}.  Our evaluation results reported in \cite{Kadkhodamohammadi2017-tx} show drastic drops in the performance of state-of-the-art methods on challenging OR data.

\par 
The availability of large-scale annotated datasets has been key to the recent progress in human detection and pose estimation.
2D datasets such as Microsoft COCO\cite{Lin2014-zn} and MPII\cite{Andriluka2014-yk} include scenes with a wide amount of variability.
3D multi-view datasets, such as Human3.6M\cite{Ionescu2014-od}, HumanEva\cite{Sigal2009-fo}, and IXMAS\cite{Weinland2010-od}, provide both 2D and 3D annotations, but with a single person performing actions in a controlled environment. As these 3D datasets do not cover real-world challenges, models trained on such data do not generalize well to these complex scenes. The MultiHumanOR dataset introduced in \cite{Belagiannis2016-dx} is a multi-view OR dataset with 2D and 3D human poses. However, it was captured during activities simulated by actors. 

We wish to introduce a new multi-view dataset illustrating the complexity of the challenging OR environment and containing data captured during real interventions. It is the first public dataset for human pose estimation featuring real clinical data and also the first multi-person multi-view RGB-D dataset containing 3D poses captured during real activities. We hope that releasing this dataset will foster research in 2D and 3D human pose estimation and the development of the related clinical applications. 

We present below the dataset, the annotations and the results of several baselines. Since some parts of the images needed to be blurred for anonymization, we also present a comparative study illustrating the impact of the blurring process on the results.

\section{MVOR Dataset}

\subsection{Data}

\begin{figure}[tb]
    \centering
    \includegraphics[width=0.6\textwidth]{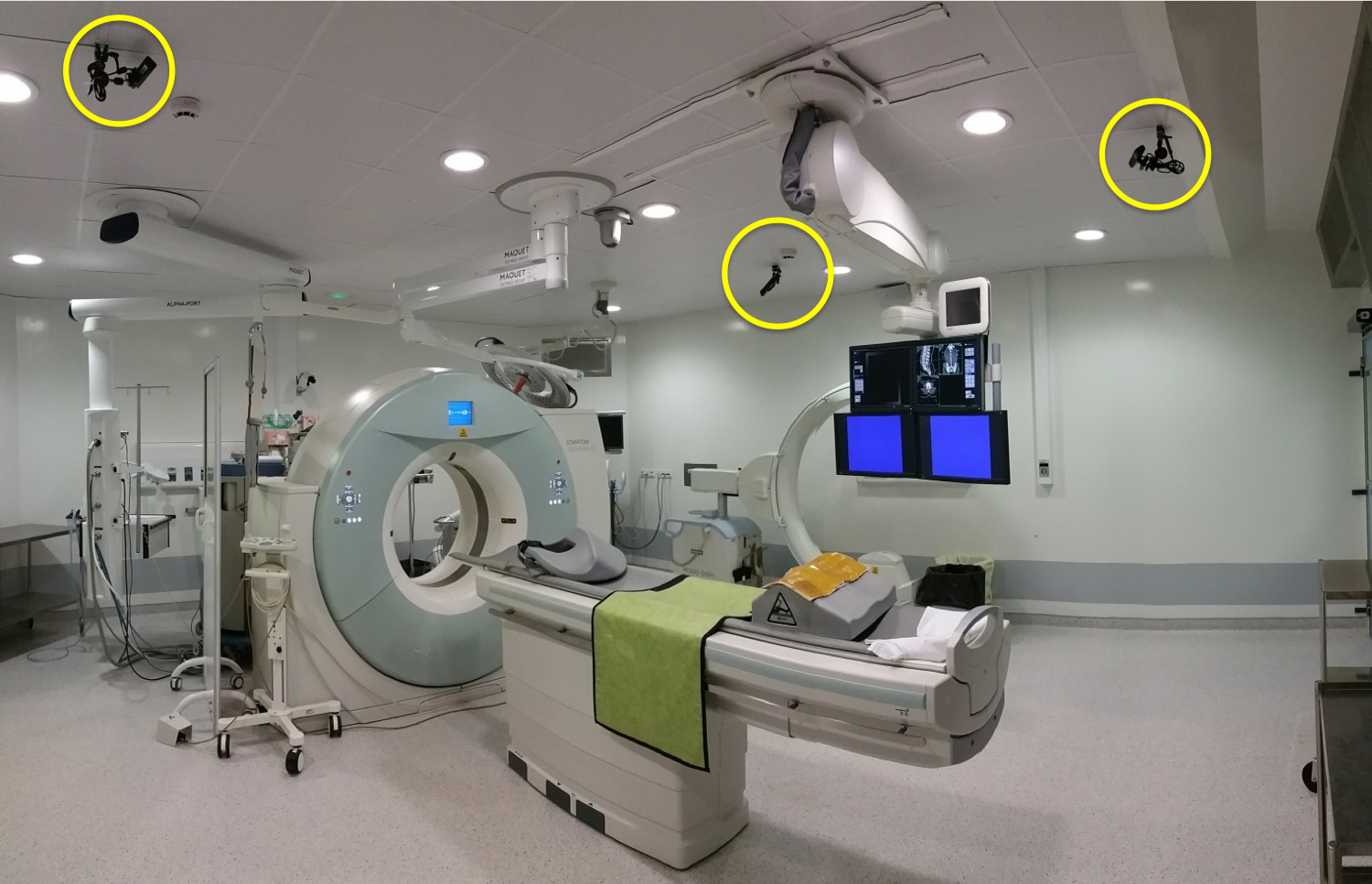}
    \caption{Multi-view setup in a room from the Interventional Radiology Department at the University Hospital of Strasbourg.}
    \label{fig:multi-cam-setup}
\end{figure}

The MVOR dataset consists of 732 multi-view frames\footnote{We define a multi-view frame as the set of RGB-D images recorded from all cameras at the same time step.} sampled from four days of recording in an interventional room at the University Hospital of Strasbourg during procedures such as vertebroplasty and lung biopsy. The ground truth annotations consist of 4699 human bounding boxes, 2926 2D upper-body poses and  1061 3D upper-body poses. The multi-view recording was performed using three RGB-D cameras (Asus Xtion Pro) mounted on the ceiling using articulated arms. The cameras were mounted in such a way as to capture the key activities around the operating table as shown in Fig. \ref{fig:multi-cam-setup}. Multi-view person and part visibility statistics are shown for the 3D annotations in Table \ref{table:stat3d}.  The image and depth data were captured at 20 FPS in 640x480 VGA resolution using a recording software developed in-house. The intrinsic camera parameters of each camera were computed using a calibration pattern. The rigid transformation between the cameras and transformation of each camera to the global coordinate system were done in the two-step process as described in \cite{Svoboda2005-jm,Loy_Rodas2015-uw}. The operating table was considered to be the reference for the global coordinate system.

\begin{figure}[tb]
    \centering
    \includegraphics[width=0.543\textwidth]{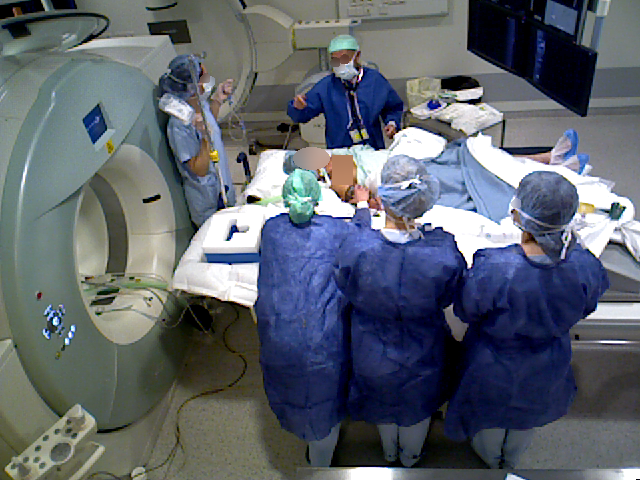}
    \hspace{0.5cm}
    \includegraphics[width=0.3\textwidth]{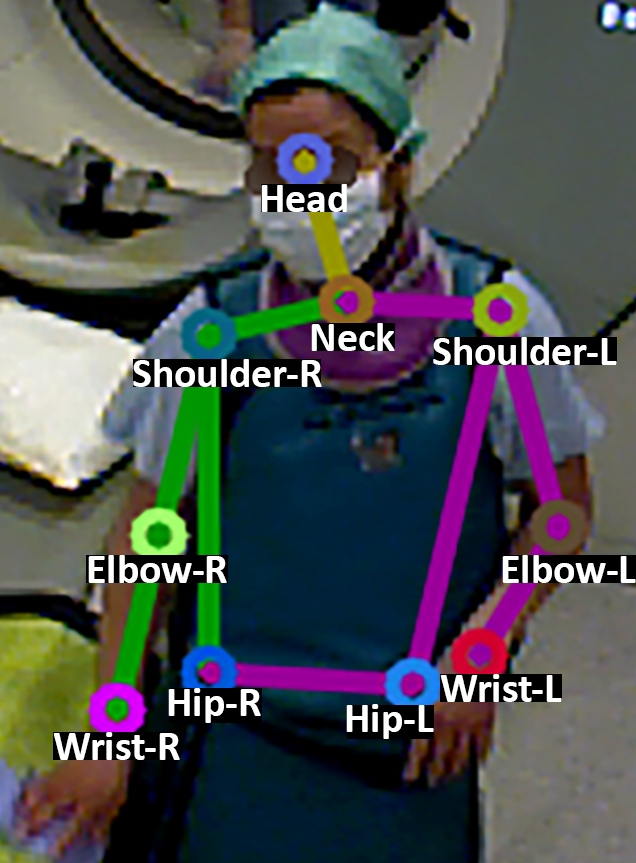}
    \caption{(Left) Illustration of the blurring process. The face of the patient, nudity and the eyes of the staff have been blurred. (Right) Keypoints of the skeleton used in the 2D/3D pose annotations.}
    \label{fig:blurring-process_skeleton}
\end{figure}

In the dataset, the color images needed to be blurred to ensure the anonymization of the data. We have tried to minimize the amount of blurring so that computer vision algorithms would not be impacted. Patient faces and nude parts are fully blurred, while clinicians' faces are only blurred around the eyes when wearing a mask, and fully blurred otherwise. A sample image is shown in Fig. \ref{fig:blurring-process_skeleton} (left).

\begin{table}[tb]
   \begin{center}
 \begin{tabular}{| l | C{1.3cm} ||  C{0.7cm} | C{0.7cm} | C{0.5cm} | C{0.5cm}| C{0.5cm} | C{0.5cm} | C{0.5cm} | C{0.5cm} | C{0.5cm} | C{0.5cm} |} 
\hline
   &Persons & Head & Neck  & \multicolumn{2}{c|}{Shoulder} & \multicolumn{2}{c|}{Hip} & \multicolumn{2}{c|}{Elbow}  & \multicolumn{2}{c|}{Wrist} \\ \cline{5-12} 
   &  & &  & L & R & L & R & L & R & L & R \\ \hline   
Three-view & 503&495& 497& 418& 464& 395& 419& 320& 391& 260& 299 \\
 \hline
Two-view & 426&419& 424& 385& 392& 354& 354& 278& 294& 191& 205\\
 \hline
One-view& 132 &127& 129& 125& 128& 119& 125& 86&  96&  55&  60\\
 \hline
\end{tabular}
\end{center}
    \caption{Multi-view statistics for the 3D annotations: number of persons and body parts visible in one, two or three views. }
\label{table:stat3d}
\end{table}

\subsection{Annotations}

\begin{figure}[t]
    \centering
    \includegraphics[width=0.9\textwidth]{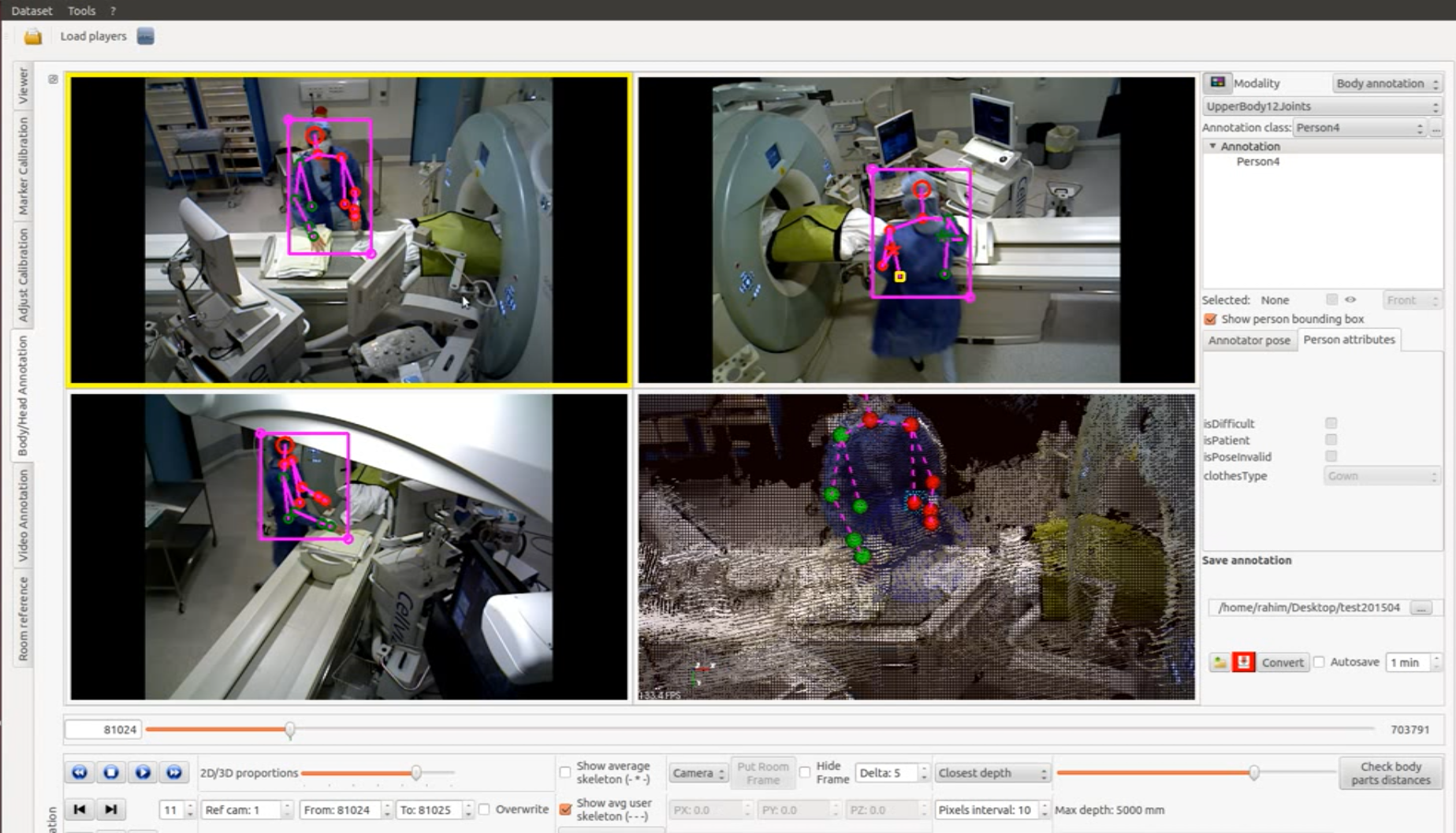}
    \caption{The tool used to generate the annotations, displaying the three views and the 3D point cloud in the interface. Right side body parts are shown in green and occluded body parts are marked by crosses. The annotators can move the joints in either 2D or 3D.}
    \label{fig:annotation tool}
\end{figure}

This dataset contains annotations for clinical staff and patients. All persons are annotated with a full bounding box and staff who have more than 50\% of their upper-body parts visible in at least one view are annotated with 2D and 3D upper-body pose keypoints. The 10 keypoints annotating the upper-body poses are shown in Fig. \ref{fig:blurring-process_skeleton} (right).

\begin{figure*}[tb]
    \centering
    \begin{subfigure}[b]{0.475\textwidth}
        \centering
        \includegraphics[width=\textwidth]{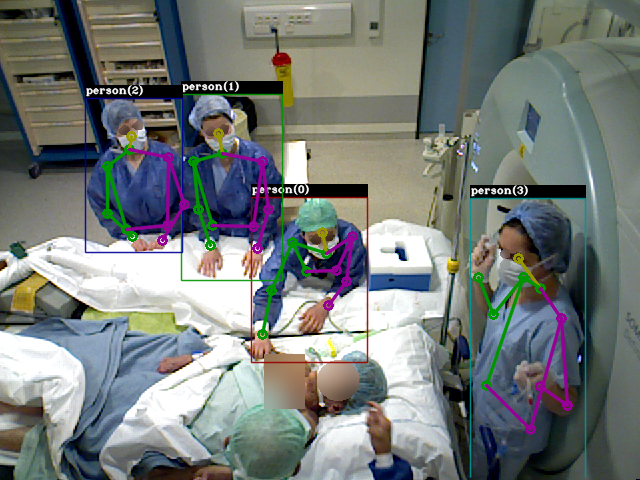}
        \caption[]%
        {{\small View 1}}    
        \label{fig:view-viz1}
    \end{subfigure}
    \hfill
    \begin{subfigure}[b]{0.475\textwidth}  
        \centering 
        \includegraphics[width=\textwidth]{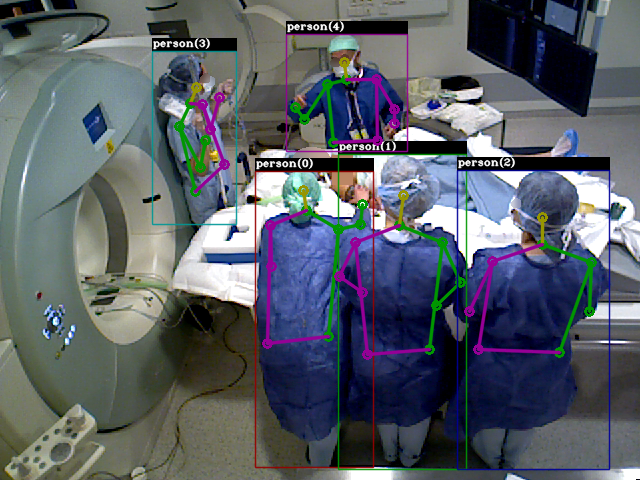}
        \caption[]%
        {{\small View 2}}    
        \label{fig:view-viz2}
    \end{subfigure}
    \vskip\baselineskip
    \begin{subfigure}[b]{0.475\textwidth}   
        \centering 
        \includegraphics[width=\textwidth]{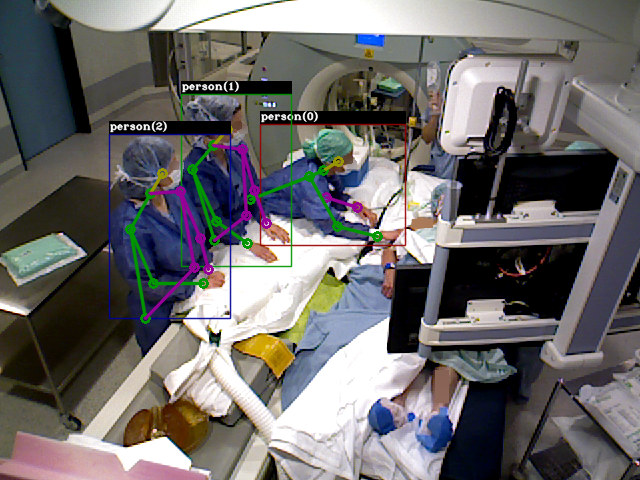}
        \caption[]%
        {{\small View 3}}    
        \label{fig:view-viz3}
    \end{subfigure}
    \quad
    \begin{subfigure}[b]{0.475\textwidth}   
        \centering 
        \includegraphics[width=\textwidth]{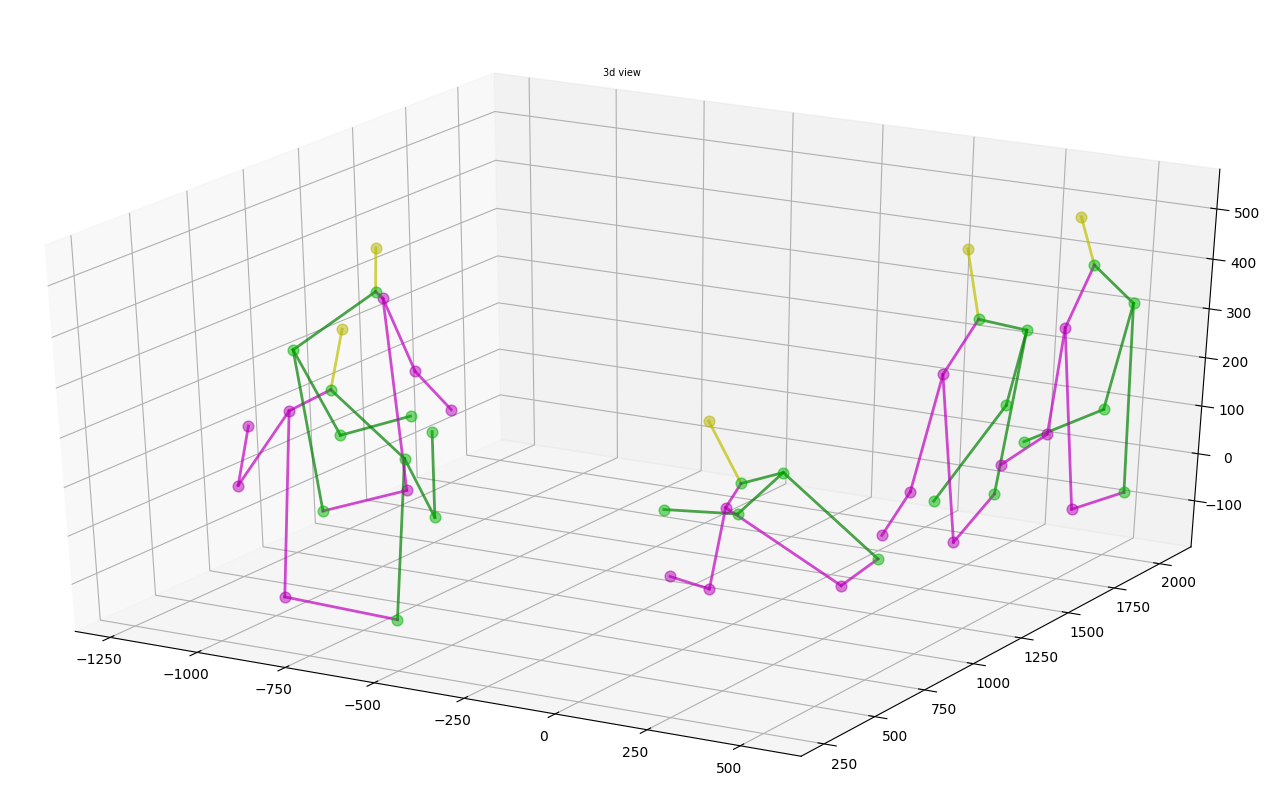}
        \caption[]%
        {{\small Virtual 3D view}}    
        \label{fig:view-3dviz}
    \end{subfigure}
    \caption[]
    {\small Visualization of 2D and 3D ground truth from the MVOR dataset.}
    \label{fig:ground_truth_viz}
\end{figure*}

To generate the annotations, we use a tool that displays all the three 2D views as well as the 3D point cloud, illustrated in Fig. \ref{fig:annotation tool}. First, the annotator draws the 2D skeletons in all 2D views. To generate the 3D annotations, the 2D poses are back-projected into 3D using the depth information and initial 3D skeletons are computed by averaging all 3D skeletons across all views. We compute average 3D joint locations only among visible body joints. These initial 3D skeletons are not always accurate due to depth errors and differences in 2D joint annotations among the views, which are in turn caused by the large visual differences due to cameras rotation angles and partial occlusions. The annotator is therefore required to then ensure the quality of each 3D skeleton by verifying the accuracy of its projections to all views and by updating its locations directly in 3D when needed. Examples of available 2D/3D annotations are shown in Figure \ref{fig:ground_truth_viz}.

\section{Baseline methods}

We briefly present the methods that we have evaluated on the MVOR dataset for 2D/3D pose estimation and for person detection.

\subsection{2D pose estimation}

{\noindent\bf Deep3DPS\cite{Kadkhodamohammadi2017-tx}:} This is a part-based approach which relies on both color and depth images to perform human pose estimation. First, a deep convolutional neural network is used to detect body parts. Then, part detection score maps and depth information are used to perform exact and efficient inference in 3D. This approach is finetuned on another RGB-D dataset captured in the same OR.
\medskip

{\noindent\bf OpenPose\cite{Cao2017-oj}:} This is an open-source\footnote[2]{The public implementation used in the evaluation is available at \href{https://github.com/CMU-Perceptual-Computing-Lab/openpose}{https://github.com/CMU-Perceptual-Computing-Lab/openpose}} bottom-up method, particularly well suited for real-time detections in RGB images. A deep multi-stage and two-branch CNN jointly predict heatmaps and part affinity fields to capture body parts and pairwise dependencies between body joints. Keypoints are then assembled into skeletons through a bipartite graph matching algorithm. For the sake of comparison, we test both OpenPose with and without multi-scale testing. We also compare to the RTPose implementation\footnote[6]{The public implementation used in the evaluation is available at \href{https://github.com/CMU-Perceptual-Computing-Lab/caffe\_rtpose}{https://github.com/CMU-Perceptual-Computing-Lab/caffe\_rtpose}}, the previous software version of OpenPose, which shows better results. 
\medskip

{\noindent\bf AlphaPose\cite{Lu2017-qr}:} This is an open-source\footnote[3]{The public implementation used in the evaluation is available at \href{https://github.com/MVIG-SJTU/AlphaPose}{https://github.com/MVIG-SJTU/AlphaPose}} top-down method for 2D pose estimation in RGB images. It performs human detection with Faster-RCNN and single person pose estimation on the extracted bounding boxes. 

\subsection{3D pose estimation}

{\noindent\bf MV3DReg\cite{Kadkhodamohammadi2018-tf}:} This approach formulates the problem of human pose estimation from multiple RGB views in a two-step framework. The method first detects 2D poses in each view independently and then relies on a neural network to regress for 3D poses. In order to easily generalize to a new environment, the approach models the characteristics of the 2D detector used at test time and proposes to only rely on a set of plausible 3D poses to learn the 3D regression function. 

\subsection{Person detection}
We evaluate person detection from RGB images with state-of-the-art convolutional object detectors, using only the detections corresponding to the person category, and with 2D pose estimators by extracting bounding boxes from predicted skeletons. 
\medskip

{\noindent\bf Faster R-CNN\cite{Ren2015-cj}:} This is the third object detector of the R-CNN family\footnote[4]{The AlphaPose implementation used in the evaluation is available at \href{https://github.com/MVIG-SJTU/AlphaPose}{https://github.com/MVIG-SJTU/AlphaPose}}. It enhances the R-CNN framework by making the region proposal network fully convolutional. Deep features are used instead of the input image to select the region of interests with a sliding window approach. Then, a second network classifies and refines the bounding box for each region of interest. 
\medskip

{\noindent\bf {Deformable convolutional networks on R-fcn\cite{Dai-2017dc}:}} This approach\footnote[5]{The public implementation used in the evaluation is available at \href{https://github.com/msracver/Deformable-ConvNets}{https://github.com/msracver/Deformable-ConvNets}} introduces a new kind of convolutional layer that extends the sampling grid of each kernel by learning an offset for each weight. These layers are integrated into the R-fcn architecture\cite{NIPS2016_6465}, a fully convolutional two-stage object detector.   
\medskip

{\noindent\bf{OpenPose\cite{Cao2017-oj} and AlphaPose\cite{Lu2017-qr}, from keypoints}:} We first use the previously described OpenPose and AlphaPose approaches to detect the poses of the persons. We then compute bounding box detections by fitting a tight bounding box around each detected skeleton.    

\section{Results}
In this section, we evaluate the state of the art methods described in the previous section for person detection and 2D/3D human pose estimation on the MVOR dataset. We also show a comparative study illustrating how these methods are affected by the blurring process. In the tables, we refer to the original (i.e, non-blurred) images with O and to the blurred images with B. 
\subsection{2D pose estimation}
We use the percentage of correct keypoints (PCK)\cite{yang2013articulated} to compare the baseline pose estimation methods. This metric measures the localization accuracy of the body joint, based on the scale of the person. To match detected and ground-truth skeletons, a tight bounding box is computed for each ground-truth skeleton from its keypoints. Then, for each ground-truth skeleton, we select the detection with the highest confidence score among the detections which have more than 30\% of their keypoints in the ground-truth bounding-box. 

Table \ref{tab:pose_estimation_2d} shows the results for AlphaPose, Deep3DPS and three versions of OpenPose. None of the 2d pose estimation algorithm is affected by the blurring process.

\begin{table}[tb]
\begin{center}
\begin{tabular}{|l|l|l|l|l|l|l|l|l|l|l|l|l|l|l|l}
\hline

\multirow{2}{*}{}             & \multicolumn{2}{c|}{Head} & \multicolumn{2}{c|}{Shoulder}  & \multicolumn{2}{c|}{Elbow} & \multicolumn{2}{c|}{Wrist} & \multicolumn{2}{c|}{Hip} & \multicolumn{2}{c|}{Average}\\ \cline{2-13} 
   & \multicolumn{1}{c|}{O} & \multicolumn{1}{c|}{B} & \multicolumn{1}{c|}{O} & \multicolumn{1}{c|}{B} & \multicolumn{1}{c|}{O} & \multicolumn{1}{c|}{B} & \multicolumn{1}{c|}{O} & \multicolumn{1}{c|}{B} & \multicolumn{1}{c|}{O} & \multicolumn{1}{c|}{B} & \multicolumn{1}{c|}{O} & \multicolumn{1}{c|}{B}    \\ \hline
Deep3DPS &  93.4 & 93.5 & 77.0 & 77.0 & 71.5 & 71.6 & 66.7 & 66.8 & 69.6 & 69.8 & 75.6 & 75.8 \\ 
\Xhline{4\arrayrulewidth}
RTPose & 91.1 & 91.0 & 88.6 & 88.8 & 74.2 & 74.5 & 57.8 & 58.1 & 56.7 & 56.4 & 73.7 & 73.8 \\
\hline
OpenPose (default) & 70.4 & 70.4 & 69.5 & 69.9 & 57.3 & 57.6 & 44.7 & 45.3 & 42.9 & 42.3 & 57.0 & 57.1 \\
\hline
OpenPose (multi-scale) & 71.0 & 71.2 & 70.3 & 70.6 & 59.7 & 60.2 & 47.6 & 47.9 & 41.9 & 41.6 & 58.1 & 58.3 \\
\Xhline{4\arrayrulewidth}
AlphaPose & 87.5 & 87.7 & 88.4 & 88.9 & 77.5 & 77.8 & 64.6 & 64.7 & 60.9 & 61.8 & 75.8 & 76.2 \\
\hline
\end{tabular}
\end{center}
\caption{PCK results for Deep3DPS, OpenPose and AlphaPose on blurred (B) and original non-blurred (O) images. }
\label{tab:pose_estimation_2d}
\end{table}

\subsection{3D pose estimation}

\begin{table}[tb]
   \begin{center}
 \begin{tabular}{|l | C{0.8cm} | C{0.8cm} | C{0.8cm} | C{0.8cm}| C{0.8cm} | C{0.8cm} | C{0.8cm} | C{0.8cm} | C{0.8cm} | C{0.8cm} |} 
\hline

\multirow{2}{*}{}   & \multicolumn{2}{c|}{Shoulder} & \multicolumn{2}{c|}{Hip} & \multicolumn{2}{c|}{Elbow}  & \multicolumn{2}{c|}{Wrist} &  \multicolumn{2}{c|}{Average} \\ \cline{2-11} 
   & O & B & O & B & O & B & O & B & O & B \\ \hline   
   
One-view &14.6 & 14.4 & 30.4 & 29.9 & 27.2 & 27.3 & 35.6 & 36.1 & 27.0 & 26.9 \\
 \hline
Two-view &8.0  & 8.1  & 16.3 & 16.0 & 19.4 & 19.4 & 29.7 & 29.8 & 18.3 & 18.3\\
 \hline
Three-view  &4.9  & 4.9  & 10.0 & 9.9  & 10.6 & 10.5 & 14.4 & 14.3 & 10.0 & 9.9\\
 \hline
\end{tabular}
\end{center}
    \caption{3D MPJP error in cm for MV3DReg on blurred (B) and original non-blurred images (O).}
\label{table:3dRes}
\end{table}

The evaluation results of MV3DReg are presented in Table \ref{table:3dRes} using the 3D mean per joint position error (MPJPE) in centimeters. We present the results per number of supporting views. The approach performs similarly on both blurred and non-blurred images, which is on par with 2D detection results\footnote{Note that MV3DReg is using 2D detections from the OpenPose implementation.}. This confirms the hypothesis that the blurring process has a negligible effect on the performance of pose estimation approaches. These results also indicate that the localization error decreases as the number of supporting view increases. This demonstrates the great benefit of multi-view data for localizing body parts in cluttered environments. 

\subsection{Person detection}
For person detection from object detectors, we use the standard metrics from COCO\cite{Lin2014-zn}, i.e average precision (AP), along with average recall ($\textrm{AR}^n$) for a fixed number of $n$ detections. $\textrm{AP}^{\textrm{\tiny IoU}}$ is the average precision for a fixed intersection over union (IoU). AP is the average of $\textrm{AP}^{\textrm{\tiny IoU}}$ for IoU from 0.50 to 0.95, with a step size of 0.05. $\textrm{AR}^{\textrm{\tiny max\_det}}$ (with $\textrm{max\_det} = [1,10,100]$) is the average of $\textrm{AR}^{\textrm{\tiny max\_det}}_{\textrm{\tiny IoU}}$ for IoU from 0.50 to 0.95, with a step size of 0.05.

For person detection from pose estimation, we choose to only report $\textrm{AP}^{50}$, because the extracted boxes are inherently smaller than the ground-truth and cannot intersect them completely, which leads to low $\textrm{AP}^{\textrm{\tiny IoU}}$ for $\textrm{IoU}>0.5$. In the same manner, we only compute AR for an IoU of $0.5$. 

Again, the difference between the results on blurred and original images is very little. As expected, the average recall is always greater on original images. However, there is sometimes a better average precision on blurred images, since there are fewer detections due to blurring.

\begin{table}[tb]
   
\begin{center}
 \begin{tabular}{|l | c | c | c | c | c | c | c | c | c | c | c | c |} 
\hline
\multirow{2}{*}{} & \multicolumn{2}{c|}{AP} & \multicolumn{2}{c|}{AP$^{50}$} & \multicolumn{2}{c|}{AP$^{75}$} & \multicolumn{2}{c|}{AR$^{1}$} & \multicolumn{2}{c|}{AR$^{10}$} & \multicolumn{2}{c|}{AR$^{100}$} \\ \cline{2-13} 
& O   & B & O   & B & O   & B & O   & B & O   & B & O & B \\ \hline 
 
 RFCN-DCN  &  0.414 & 0.404 & 0.640 & 0.620 & 0.473 & 0.466 & 0.274 & 0.271 & 0.462 & 0.451 & 0.462 & 0.451\\ 
 \hline

 Faster-RCNN  & 0.523 & 0.504 & 0.802 & 0.756 & 0.586 & 0.574 & 0.303 & 0.300 & 0.625 & 0.598 & 0.630 & 0.601\\
 \hline
\end{tabular}
\end{center}
  
 \caption{AP and AR results for person detection  from two state-of-the-art methods: Faster-RCNN\cite{Ren2015-cj} and Deformable Convolutional Nets\cite{Dai-2017dc} on R-FCN\cite{NIPS2016_6465}. }
\end{table}

\begin{table}[tb]
\begin{center}
\begin{tabular}{|l|c|c|c|c|c|c|}
\hline
\multirow{2}{*}{}              & \multicolumn{2}{c|}{AP$^{50}$} & \multicolumn{2}{c|}{AR$^{1}_{\textrm{\tiny IoU} = 0.5}$} & \multicolumn{2}{c|}{AR$^{10}_{\textrm{\tiny IoU} = 0.5}$} \\ \cline{2-7} 
                               & O   & B & O   & B & O   & B \\ \hline
RTPose                         & 0.364 & 0.371 & 0.272 & 0.274 & 0.480 & 0.475    \\ 
\Xhline{4\arrayrulewidth}
OpenPose (default)             & 0.278 & 0.278 & 0.303 & 0.299 & 0.416 & 0.406 \\ \hline
OpenPose (multi-scale)         & 0.268 & 0.267 & 0.291 & 0.289 & 0.400 & 0.394 \\ 
\Xhline{4\arrayrulewidth}
AlphaPose                      & 0.540 & 0.512 & 0.338 & 0.331 & 0.676 & 0.636 \\ 
\hline
\end{tabular}
\end{center}
\caption{$\textrm{AP}^{50}$ and AR results for person detection using bounding boxes generated from the detected human poses. }
\label{Person_detection_from_keypoints}
\end{table}

\section{Conclusions}
In this paper, we present and release a new multi-view dataset for multi-person detection and 2D/3D human pose estimation in a challenging environment, namely a modern operating room, which contains inherent visual challenges such as multiple occlusions. We also present the results of several recent baseline methods as well as a comparative study showing that the blurring process required for medical confidentiality only affects mildly the accuracy of detections. This dataset can thus be helpful to evaluate a detector's ability to generalize to unseen configurations and color distribution and also to assess the performance of 3D multi-person pose estimation methods on real-world data.

\subsubsection{Acknowledgements.}
This work was supported by French state funds managed by the ANR within the Investissements d'Avenir program under references ANR-16-CE33-0009 (DeepSurg), ANR-11-LABX-0004 (Labex CAMI) and ANR-10-IDEX-0002-02 (IdEx Unistra). The authors would also like to thank the members of the Interventional Radiology Department at University Hospital of Strasbourg for their help in generating the dataset.

\bibliographystyle{splncs03}
\bibliography{references}

\end{document}